\documentclass{article}

\usepackage{arxiv}

\usepackage[utf8]{inputenc} %
\usepackage[T1]{fontenc}    %
\usepackage{hyperref}       %
\usepackage{url}            %
\usepackage{booktabs}       %
\usepackage{amsfonts}       %
\usepackage{nicefrac}       %
\usepackage{microtype}      %
\usepackage{lipsum}		%
\usepackage{graphicx}
\usepackage{natbib}
\usepackage{doi}

\usepackage{xcolor}
\usepackage{caption}
\usepackage{subcaption}     %
\usepackage{amsmath}
\usepackage{lscape}

\title{Quality Assessment of Photoplethysmography Signals For Cardiovascular Biomarkers Monitoring Using Wearable Devices}

\author{Felipe M.~Dias \\
	Heart Institute (InCor)\\
	Clinics Hospital University of \\
 Sao Paulo Medical School\\
	Sao Paulo - SP - Brazil \\
	\texttt{f.dias@hc.fm.usp} \\
 \And
	Marcelo A.~F.~Toledo \\
	Heart Institute (InCor)\\
	Clinics Hospital University of \\
 Sao Paulo Medical School\\
	Sao Paulo - SP - Brazil \\
	\texttt{marcelo.arruda@hc.fm.usp.br} \\
 \And
	Diego A.~C.~Cardenas \\
	Heart Institute (InCor)\\
	Clinics Hospital University of \\
 Sao Paulo Medical School\\
	Sao Paulo - SP - Brazil \\
	\texttt{diego.cardona@hc.fm.usp.br} \\
 \And
 Douglas A.~Almeida \\
	Heart Institute (InCor)\\
	Clinics Hospital University of \\
 Sao Paulo Medical School\\
	Sao Paulo - SP - Brazil \\
	\texttt{douglas.andrade@hc.fm.usp.br} \\
 \And
	Filipe A.~C.~Oliveira \\
	Heart Institute (InCor)\\
	Clinics Hospital University of \\
 Sao Paulo Medical School\\
	Sao Paulo - SP - Brazil \\
	\texttt{filipe.acoliveira@hc.fm.usp.br} \\
 \And
	Estela ~Ribeiro \\
	Heart Institute (InCor)\\
	Clinics Hospital University of \\
 Sao Paulo Medical School\\
	Sao Paulo - SP - Brazil \\
	\texttt{estela.ribeiro@hc.fm.usp.br} \\
 \And
	Jose E.~Krieger \\
	Heart Institute (InCor)\\
	Clinics Hospital University of \\
 Sao Paulo Medical School\\
	Sao Paulo - SP - Brazil \\
	\texttt{j.krieger@hc.fm.usp.br} \\
  \And
	Marco A.~Gutierrez \\
	Heart Institute (InCor)\\
	Clinics Hospital University of \\
 Sao Paulo Medical School\\
	Sao Paulo - SP - Brazil \\
	\texttt{marco.gutierrez@incor.usp.br}
}

\date{}

\renewcommand{\shorttitle}{Quality Assessment Of PPG signals}

\hypersetup{
pdftitle={A template for the arxiv style},
pdfsubject={q-bio.NC, q-bio.QM},
pdfauthor={David S.~Hippocampus, Elias D.~Striatum},
pdfkeywords={First keyword, Second keyword, More},
}

\begin{document}
\maketitle

\begin{abstract}
Photoplethysmography (PPG) is a non-invasive technology that measures changes in blood volume in the microvascular bed of tissue. It is commonly used in medical devices such as pulse oximeters and wrist-worn heart rate monitors to monitor cardiovascular hemodynamics. PPG allows for the assessment of parameters (e.g., heart rate, pulse waveform, and peripheral perfusion) that can indicate conditions such as vasoconstriction or vasodilation, and provides information about microvascular blood flow, making it a valuable tool for monitoring cardiovascular health. However, PPG is subject to a number of sources of variations that can impact its accuracy and reliability, especially when using a wearable device for continuous monitoring, such as motion artifacts, skin pigmentation, and vasomotion. In this study, we extracted 27 statistical features from the PPG signal for training machine-learning models based on gradient boosting (XGBoost and CatBoost) and Random Forest (RF) algorithms to assess quality of PPG signals that were labeled as good or poor quality. We used the PPG time series from a publicly available dataset and evaluated the algorithm's performance using Sensitivity (Se), Positive Predicted Value (PPV), and F1-score (F1) metrics. Our model achieved Se, PPV, and F1-score of 94.4\%, 95.6\%, and 95.0\% for XGBoost, 94.7\%, 95.9\%, and 95.3\% for CatBoost, and 93.7\%, 91.3\% and 92.5\% for RF, respectively. Our findings are comparable to state-of-the-art reported in the literature but using a much simpler model, indicating that ML models are promising for developing remote, non-invasive, and continuous measurement devices.
\end{abstract}


\section{Introduction}
Photoplethysmography (PPG) is a non-invasive technology that measures changes in blood volume in the microvascular bed of tissue, and is widely used in pulse oximetry (SpO2) devices to assess cardiovascular health \cite{alian2014photoplethysmography} \cite{REISNER2008}.
In addition to pulse oximetry, PPG can potentially be used to measure other important parameters such as heart rate, respiratory rate, and other physiological parameters over time \cite{charlton2022wearable}. 
With its non-invasive nature and ability to provide continuous monitoring, PPG has become an important tool for monitoring cardiovascular health and diagnosing various cardiovascular conditions \cite{MEJIAMEJIA2022}. 

Generally speaking, PPG measures changes in the blood volume of vascular tissues by shining light over a peripheral tissue and measuring the amount of light that is absorbed and scattered. The attenuated light is detected by an optical sensor, which records the changes in light intensity over time \cite{MEJIAMEJIA2022} \cite{NITZAN2022}. 
The PPG signal consists of an  Alternating Current (AC) component, which is the high-intensity component caused by the light absorption of hemoglobin in pulsatile arterial blood, and a Direct Current (DC) component, which is the low intensity component caused by changes in other tissues components and nonpulsatile arterial blood \cite{MUKKAMALA2022}. The AC component of the PPG signal reflects the changes in blood volume due to the oscillations in blood cell aggregation and blood flow related to changes in arterial blood pressure. Specifically, during systole, when the heart pumps blood and arterial blood pressure increases, there is an increased absorbance of light, leading to a higher AC component of the PPG signal. During diastole, when the heart is filling with blood and arterial blood pressure decreases, there is a decreased absorbance of light, leading to a lower AC component of the PPG signal \cite{NITZAN2022}. 

\begin{figure}[!h]
	\centering
	\includegraphics[width=0.55\linewidth]{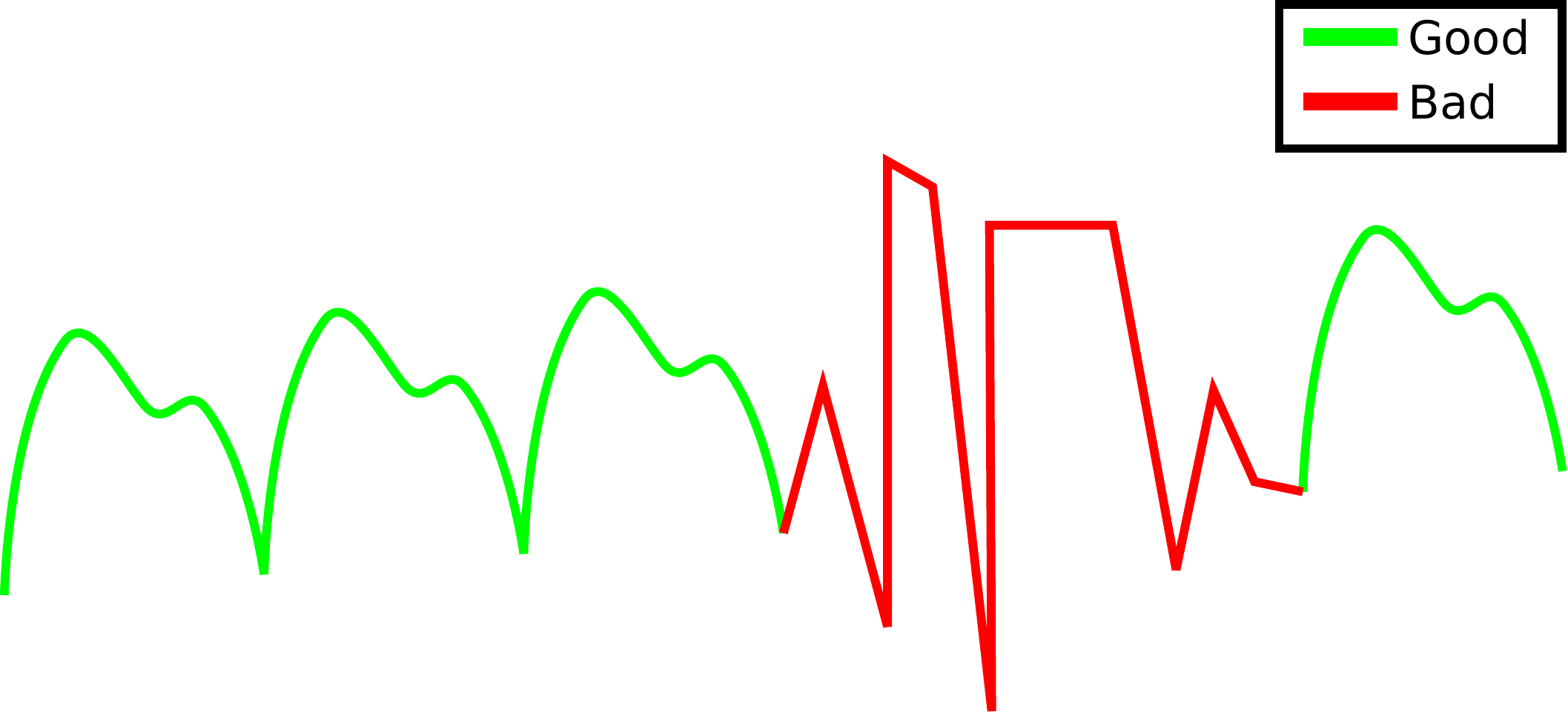}
	\caption{Individual PPG heartbeat signal for two different quality levels: Good and Bad.}
	\label{fig: elgendi}
\end{figure}

The quality of PPG signals obtained from wearable devices is a major concern \cite{fine2021}. %
PPG signals can be affected by various sources of noise, including motion artifact, probe-tissue interface disturbance, such as pressure between the PPG sensor and the skin, baseline interference due to respiration and body movement, low and high frequency noise, and type of sensor and location of the measurement \cite{elgendi2012analysis} \cite{li2018comparison}.
In this context, \cite{MOSCATO2022A} showed that physical activity affected PPG signal quality in a way that, during rest, 94\% of the heartbeats were considered of good quality compared to only 9\% during physical activity and suggested that healthy subjects have a better signal quality (44\% of good quality) compared to oncological patients (13\% of good quality). %
These factors can negatively impact the PPG signal analysis and hinder the extraction of meaningful features and biomarkers, or even act as confounding factors, invalidating their usage or interpretation.
In Fig. \ref{fig: elgendi}, we show an illustrative example of a good PPG signal segment, where physiological measurements (e.g., heart rate) can be easily extracted, and a bad PPG signal where such measurements are not reliable.

Therefore, developing a good signal quality detector for PPG signals is crucial in order to ensure that the signals obtained from wearable devices are of high quality and suitable for analysis. 
However, the lack of labeled and publicly available datasets of signal quality assessment is a major issue, as it makes it difficult to train and validate the performance of signal quality detectors. 

One of the earliest attempts to approach signal quality in PPG signal was proposed by \cite{ELGENDI2016}. In his work, he proposed a recommendation for visual quality assessment annotation of individual heartbeats following three quality levels: (1) excellent, for PPG signals where systolic and diastolic peaks can be clearly detected; (2) acceptable, for PPG signals where heart rate can still be estimated even though the diastolic peak is not salient; and (3) unfit, for noisy PPG signals where systolic and diastolic peaks cannot be distinguished.

Recently, \cite{TORRES-SOTO2020} created a public PPG quality dataset and proposed the DeepBeat algorithm that used segments of 25 seconds of PPG signals as input to predict signal quality, along with Atrial Fibrillation (AF) rhythm classification. \cite{MOHAGHEGHIAN2022} used multiple databases to assess the quality of PPG signals by extracting statistical and morphological features from the signals. \cite{MOSCATO2022A}, using a private dataset, employed a combination of features extracted features both from the accelerometers and PPG waveform to estimate signal quality. \cite{DIAS2022} used the MIMIC-II \cite{MIMIC2} dataset and employed a template-based method to assess the quality of PPG signals. 
They detected the beats in each window of the signal, estimated a template average beat, and computed Pearson’s correlation coefficient between each beat and the template average beat. Windows whose mean correlation was lower than a given threshold were considered to have poor signal quality. The field of automatic assessment of PPG signals for cardiovascular biomarkers monitoring is still in its early stages, and there is much room for improvement and further research. 

In this study, we propose a signal quality assessment method for PPG signals by extracting 27 statistical features from 25 s segments of these signals. 
These features are fed to a machine learning model (XGBoost, CatBoost, or Random Forest) which classify the segment as either good or bad quality.
The proposed method aims to reduce the effect of noisy beats on PPG signals, improving further analysis and applications.

\section{Materials and Methods}

\begin{figure}[!t]
	\centering
	\includegraphics[width=1\linewidth]{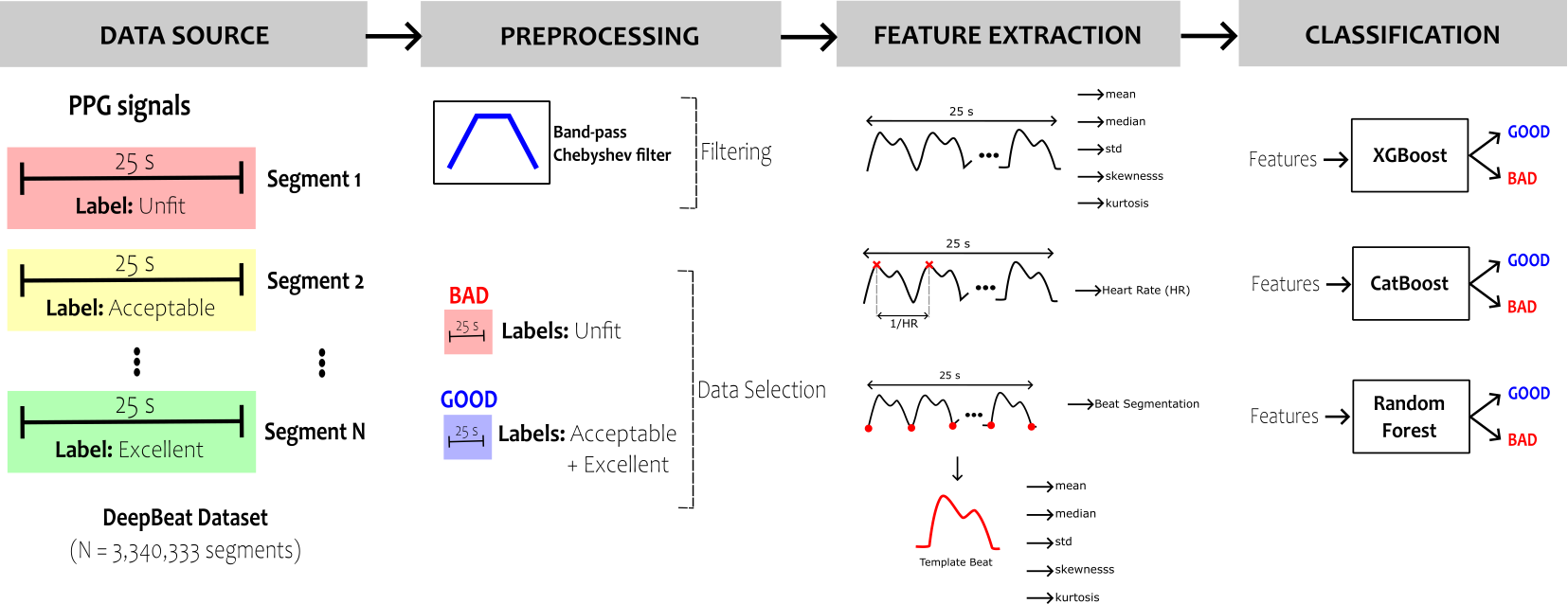}
	\caption{General structure of the proposed methodology to classify PPG signal quality.}
	\label{fig: methodology}
\end{figure}

In this section, we describe the publicly available dataset used to classify the PPG signal quality (A). Thereafter, our proposed methodology for signal quality classification is described, based on feature extraction of PPG signals (B, C) and the use of three algorithms for the classification step (D). The general structure of the proposed method to classify PPG signal quality is shown in Fig. \ref{fig: methodology}.

\subsection{Data source}

We used a publicly available dataset provided by Stanford University \cite{TORRES-SOTO2020} named DeepBeat. This dataset is composed of three different types of signals. In the first part of the dataset, they collected data using a wrist PPG wearable device (Simband), sampled at 128 Hz, from subjects with confirmed Atrial Fibrillation diagnosis, performing elective cardioversion or stress tests. For the second part of the dataset, they generated synthetic physiological signals of sinus rhythms and atrial fibrillation rhythms, adding noise components to this synthetic dataset, simulating high-quality signals with low noise or low-quality signals with high noise. The third and last part of the dataset is composed of PPG data from the 2015 IEEE Signal Processing Cup, to include signals from healthy subjects. 
The DeepBeat dataset provides signals partitioned into segments of 25 s and split into training, validation, and test partitions, avoiding data leakage, i.e., samples from the same subject don’t appear in the training, validation, and test sets. 
To provide the labels for each 25 s segments, the authors of this dataset used the Elgendi’s quality assessment \cite{ELGENDI2016} recommendation and labeled 1,000 randomly selected segments. A separate model was trained with these 1,000 labeled segments, predicting quality labels for all the remaining segments of the dataset. In summary, the DeepBeat dataset provides 25 s segments labeled by a model proposed by the dataset authors into three classes, i.e., Excellent, Acceptable and Unfit. Table \ref{tab: dataset} show the number of 25 s segments for each class, distributed on each partition, available on DeepBeat dataset.

\begin{table}[!h]
    \centering
    \caption{Summary of the DeepBeat dataset.}
    \label{tab: dataset}
    \begin{tabular}{cccc|c}
    \hline
    \textbf{Class} & \textbf{Training} & \textbf{Validation} & 
    \textbf{Test} & \textbf{Total}\\ \hline
    \textit{Excellent} & 550,702 & 124,995 & 3,246 & 678,943\\
    \textit{Acceptable} & 281,024 & 64,647 & 2,032 & 347,703\\
    \textit{Unfit} &  1,972,208 & 329,140 & 12,339 & 2,313,687\\ \hline
    Total &  2,803,934 & 518,782 & 17,617 & 3,340,333 \\ \hline
    \end{tabular}
\end{table}

With this approach, this dataset doesn’t score individual heartbeats, as proposed in Elgendi’s quality assessment \cite{ELGENDI2016}, and the authors didn’t make it clear what criteria they used to determine if a given segment was defined as excellent, acceptable, or unfit. Besides, the authors didn’t specify which dataset a particular segment corresponds to, i.e., the one collected from AF subjects, the synthetic, or the healthy subjects dataset. Furthermore, most of the data was labeled by a quality assessment model, which means that the labels are not entirely reliable, and the metrics of this model’s performance were not presented in \cite{TORRES-SOTO2020}. Regardless of these constrains, this is the only publicly available dataset on this matter, to the best of our knowledge. 

\subsection{Preprocessing}

In our work, we decided to merge the labels excellent and acceptable signals (see Table \ref{tab: dataset}) into one unique label, resulting in a binary class, i.e., Good (excellent and acceptable) and Bad signals (Unfit). 
All the PPG signal 25 s segments were filtered using a 4th-order Chebyshev type II bandpass filter of 0.5 -- 10 Hz. 

\subsection{Feature extraction}

\begin{figure}[!h]
	\centering
	\includegraphics[width=0.7\linewidth]{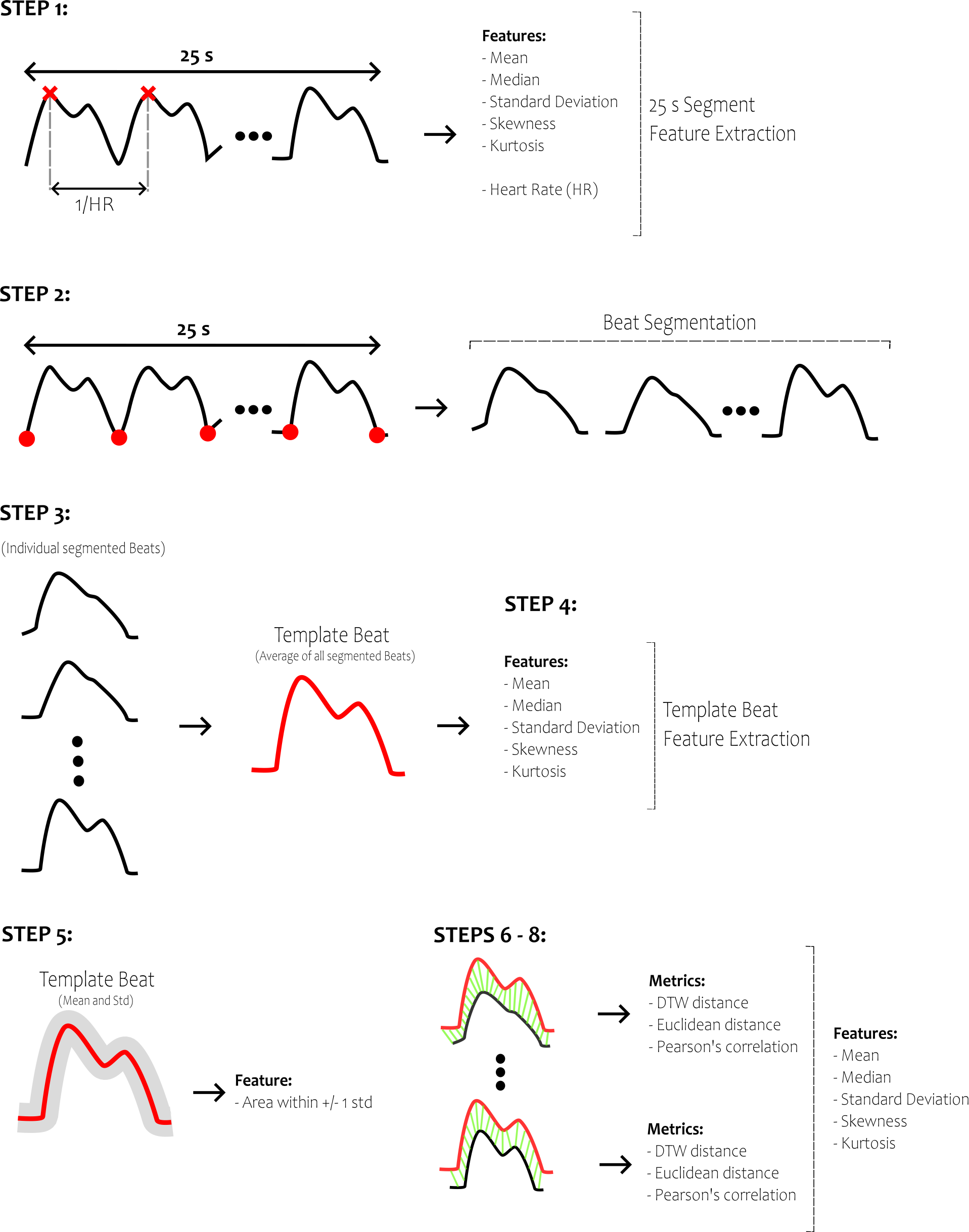}
	\caption{Overview of the 8 feature extraction steps.}
	\label{fig: preprocessing}
\end{figure}

To extract features from each of the PPG segments, we performed the following steps, as illustrated in Figure \ref{fig: preprocessing}:

\begin{enumerate}
    \item \textit{Step 1:} Extracted Heart Rate (HR) and 5 statistical features (mean, median, standard deviation, skewness, and kurtosis) from the full 25 s segments;
    \item \textit{Step 2:} Performed beat segmentation using the Multi-Scale Peak and Trough Detection (MSPTD) \cite{MQSPD} algorithm for each segment, resulting in $N$ beats per segment;
    \item \textit{Step 3:} Defined a Template Beat as the average beat of the segment;
    \item \textit{Step 4:} Extracted 5 statistical features (mean, median, standard deviation, skewness, and kurtosis) from the Template Beat;
    \item \textit{Step 5:} Computed the area within $\pm$ 1 std of the Template Beat according to the remaining beats;
    \item \textit{Step 6:} Computed the Dynamic Time Warping (DTW) distance for each individual beat with the Template Beat ($\mathbf{DTW} = [DTW_1, DTW_2, \dots, DTW_N]$) and extracted 5 statistical features from the resulting $\mathbf{DTW}$ vector;
    \item \textit{Step 7:} Computed the Euclidean distance for each individual beat with the Template Beat ($\mathbf{E} = [E_1, E_2, \dots, E_N]$) and extracted 5 statistical features from the resulting $\mathbf{E}$ vector;
    \item \textit{Step 8:} Computed Pearson's correlation for each individual beat with the Template Beat ($\mathbf{\rho} = [\rho_1, \rho_2, \dots, \rho_N]$) and extracted 5 statistical features from the resulting $\mathbf{\rho}$ vector. %
\end{enumerate}

These steps result in a set of $n = 27$ features that were used to perform the quality assessment of the PPG signals. 
Table \ref{tab: features} displays a summary of the features extracted. 

\begin{table}[!h]
\centering
\caption{Summary of the features extracted.}
\label{tab: features}
\begin{tabular}{ll}
\hline
\textbf{Index  } & \textbf{Feature name} \\ \hline
0 & Mean of the full 25 s segments \\
1 & Median of the full 25 s segments \\
2 & Standard deviation of the full 25 s segments \\
3 & Skewness of the full 25 s segments \\
4 & Kurtosis of the full 25 s segments \\
5 & Heart Rate \\
6 & Mean of the template \\
7 & Median of the template \\
8 & Standard deviation of the template \\
9 & Skewness of the template \\
10 & Kurtosis of the template \\
11 & Area within $\pm$ 1 std of the template \\
12 & Mean DTW distance \\
13 & Median DTW distance \\
14 & Standard deviation DTW distance \\
15 & Skewness DTW distance \\
16 & Kurtosis DTW distance \\
17 & Mean Euclidean distance \\
18 & Median Euclidean distance \\
19 & Standard deviation Euclidean distance \\
20 & Skewness Euclidean distance \\
21 & Kurtosis Euclidean distance \\
22 & Mean Pearson's correlation \\
23 & Median Pearson's correlation \\
24 & Standard deviation Pearson's correlation \\
25 & Skewness Pearson's correlation \\
26 & Kurtosis Pearson's correlation \\ \hline
\end{tabular}
\end{table}

\subsection{Classification (D)}
We used three traditional and well-known machine learning algorithms for the quality assessment of the PPG signals, being them: XGBoost \cite{xgboost}; CatBoost \cite{catboost}; and Random Forest \cite{randomforest}. 
XGBoost (eXtreme Gradient Boosting) \cite{xgboost} is a gradient boosting algorithm used for classification and regression tasks, capable to handle complex patterns in data, and deliver high model performance, providing efficient computation for large datasets.
Likewise, CatBoost (Categorical Boosting) \cite{catboost} is another gradient boosting algorithm designed to handle categorical features. 
Finally, Random Forest \cite{randomforest} is an ensemble learning algorithm based on the concept of decision trees,  but instead of using a single decision tree, Random Forest combines multiple decision trees to make predictions in a more robust and accurate manner, providing a balance between performance and interpretability.
For these three methods, we employed their default hyperparameter values.

The DeepBeat dataset \cite{TORRES-SOTO2020} already provide the data split in three sets: training, validation and testing, as described on Table \ref{tab: dataset}. 
Our results were generated using the testing set. To evaluate the employed models, we assessed three different metrics, including Sensitivity (Se), Positive Predicted Value (PPV), and F1-score (F1). 
Sensitivity measures the proportion of actual positive samples that are correctly predicted, positive predictive value measures the proportion of predicted positive samples that are correct, and F1-score is a single performance metric that balances sensitivity and positive predictive value.

\subsection{Experimental setup}

The experiments were performed in Python (3.8.10) with the support of the libraries scikit-learn (1.1.3), XGBoost (1.6.1), numpy (1.19.5), scipy (1.8.1), and catboost (1.1.1). 

\section{Results}

Table \ref{tab: results} provides the performance results of the three proposed models for PPG quality assessment and the comparison with the results found in the current state-of-the-art works. 
We achieved Sensitivity (Se), Positive Predictive Value (PPV), and F1-score of 94.4\%, 95.6\%, and 95.0\%, respectively, using the XGBoost method.
Additionally, using Random Forest, we obtained 93.7\%, 91.3\%, and 92.5\% for Se, PPV, and F1-score, respectively.
Our best results were obtained using CatBoost, achieving Se of 94.7\%, PPV of 95.9\%, and F1-score of 95.3\%.

\begin{table}[!h]
\centering
\caption{Comparison of the performance results of our three proposed algorithms for PPG quality assessment and the related state-of-the-art works.}
\label{tab: results}
\begin{tabular}{lcccc}
\hline
\textbf{Algorithm} & \textbf{Dataset} & \textbf{Se}(\%) & \textbf{PPV}(\%) & \textbf{F1}(\%)  \\ \hline
\textit{Our method: CatBoost} & DeepBeat (test set) & 94.7 & 95.4 & 95.0 \\                        
\textit{Our method: XGBoost}  & DeepBeat (test set) & 94.6 & 95.2 & 94.9 \\
\textit{Our method: Random Forest} & DeepBeat (test set) & 94.6 & 95.0 & 94.8 \\
\textit{\cite{TORRES-SOTO2020}} & DeepBeat (test set) & 97.6  & 97.4 & 97.5 \\
\textit{\cite{MOHAGHEGHIAN2022}} & Multiple datasets & 86.2  & 98.4 & 91.9 \\ \hline
\end{tabular}
\end{table}

\begin{figure}[h!]
\centering
\begin{subfigure}{.45\textwidth}
  \centering
  \includegraphics[width=\linewidth]{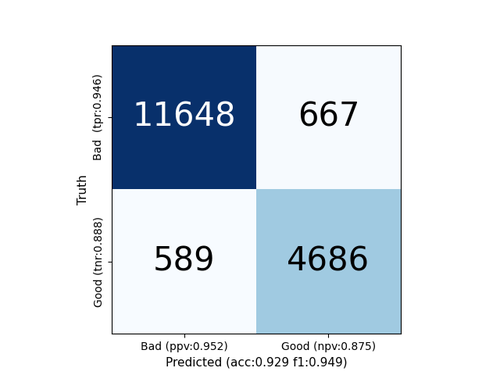}
  \caption{}
  \label{fig: confusion_matrix_xgboost}
\end{subfigure}%
\begin{subfigure}{.45\textwidth}
  \centering
  \includegraphics[width=\linewidth]{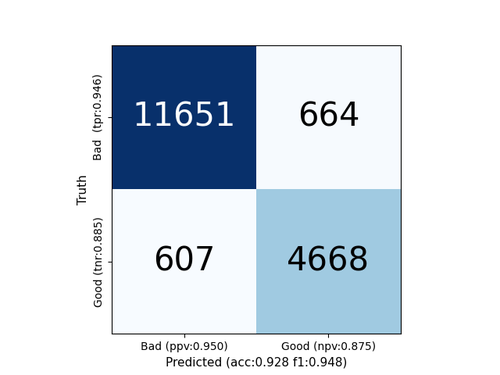}
  \caption{}
  \label{fig: confusion_matrix_rf}
\end{subfigure}
\begin{subfigure}{.45\textwidth}
  \centering
  \includegraphics[width=\linewidth]{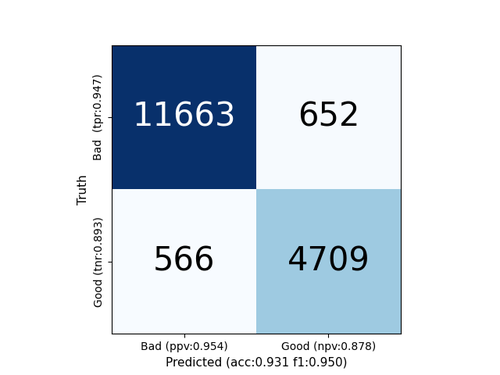}
  \caption{}
  \label{fig: confusion_matrix_catboost}
\end{subfigure}
\caption{Confusion Matrix of the three proposed algorithms on the test set: (a) XGBoost; (b) Random Forest and (c) CatBoost.}
\label{fig: confusion_matrix}
\end{figure}

Figures \ref{fig: confusion_matrix_xgboost}, \ref{fig: confusion_matrix_rf} and \ref{fig: confusion_matrix_catboost} displays the confusion matrix for the three proposed algorithms on the test set.

\begin{figure}[!t]
	\centering
	\includegraphics[width=0.9\linewidth]{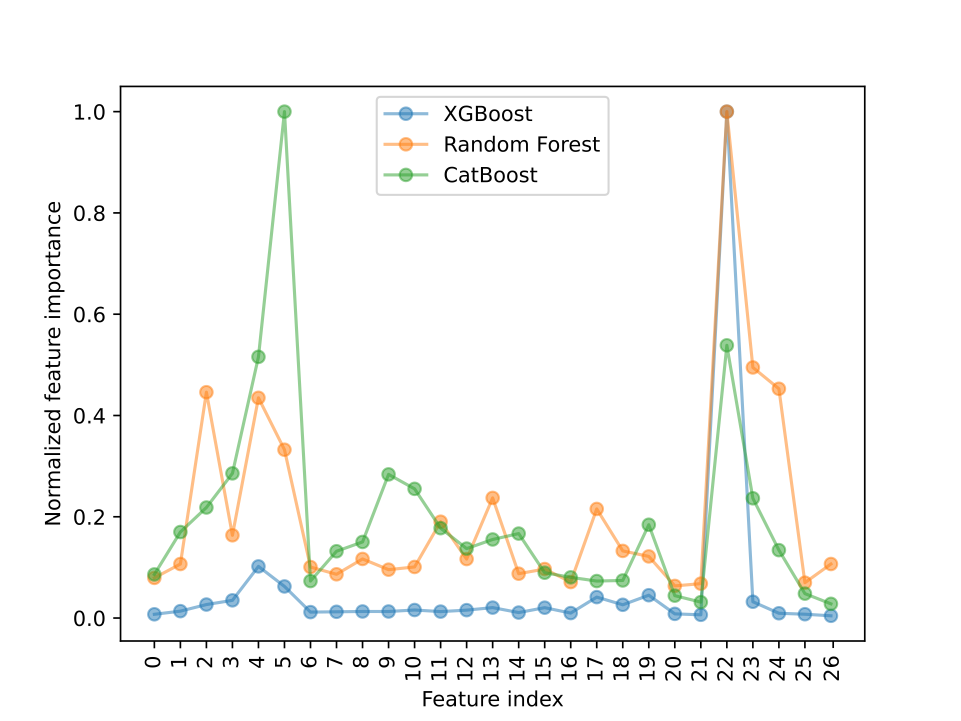}
	\caption{Feature importance measures for the three proposed algorithms.}
	\label{fig: feature_importance}
\end{figure}

Finally, we show the feature importance measures on Figure \ref{fig: feature_importance}, providing insights into which features are most important in predicting the outcome for the three proposed models.

\section{Discussion}

In this study, we propose a promising alternative method to evaluate the quality of photoplethysmography (PPG) signals by comparing it with other methodologies proposed in the literature. It is worth noting that only a few studies have proposed methodologies for assessing PPG signal quality, mostly using different datasets. Therefore, comparing our results with those of other studies is challenging.

We obtained our best results using the CatBoost algorithm, achieving Sensitivity (Se), Positive Predictive Value (PPV), and F1-score of 94.7\%, 95.9\%, and 95.3\%, respectively.
Compared to other feature-based approach, i.e., Mohagheghian et al. \cite{MOHAGHEGHIAN2022}, we obtained a superior F1-score. 
On the other hand, compared to the deep learning approach proposed by Torres-Soto and Ashley \cite{TORRES-SOTO2020}, we achieved competitive results using a simpler approach. 

Moreover, features 5 and 22, Heart Rate and Mean Pearson's correlation respectively, are indicated as the most important for our CatBoost algorithm. 
XGBoost algorithm considers the Mean Pearson's correlation feature as the most relevant. 
Random Forest algorithm also considers this feature as the most relevant. 
It seems that Mean Pearson's correlation feature carry important information on the quality of the PPG signal. Even though this feature is significant for all algorithms, Random Forest and CatBoost algorithms still consider other features on their predictions. %

Our results, which used a less complex model than the state-of-the-art, demonstrate that Machine Learning models are advantageous for creating a pipeline for assessing PPG signal quality, reducing noise and improving the reliability of subsequent analyses and applications on PPG devices used for remote, non-invasive, and continuous monitoring.

Since PPG signals can be affected by various noise sources due to motion disturbances and acquisition condition, which can reduce their morphological quality, it consequently can impact the accuracy of the information retrieved \cite{MOSCATO2022A}. 

PPG quality assessment can also help to identify faulty wearable devices and prevent false measurements. This is important in clinical settings, where accurate and reliable measurements are critical for making informed decisions about patient care. 
This is especially important for the monitoring of cardiovascular biomarkers, as inaccurate or unreliable measurements can have severe implications for the diagnosis, treatment, and management of cardiovascular diseases. 

The automatic signal quality evaluation technique proposed here aims to increase the reliability of the PPG parameters and expand its practical applicability. 
Future works on PPG quality assessment for cardiovascular biomarkers monitoring using wearable devices should consider validate the proposed methods on a larger and more diverse dataset, including patients with different medical conditions, ages, and ethnicities, to ensure their effectiveness and reliability. To do so, it is necessary the development of well-annotated datasets that include a wide range of signal qualities, including low and high-quality signals, and different types of noise sources, such as motion artifacts, ambient light, and physiological noise.

\section{Conclusion}

The proposed signal quality assessment of PPG signals can help improve the accuracy and reliability of physiological parameters, such as respiratory rate and heart rate, in wearable devices by reducing the impact of unfavorable factors such as motion disturbances. This is important in the context of continuous monitoring and to ensure the wide applicability of PPG signals in various applications. 

In conclusion, our proposed method of PPG signal quality assessment using statistical features shows promising results.
However, the limitations of the available database used in this study need to be addressed for a fairer evaluation. 
Further improvement and validation of the method is necessary with a larger and more diverse dataset, which would lead to a more robust and practical PPG signal quality assessment for various applications.

\bibliographystyle{unsrtnat}
\bibliography{references}  %

\begin{thebibliography}{19}
\providecommand{\natexlab}[1]{#1}
\providecommand{\url}[1]{\texttt{#1}}
\expandafter\ifx\csname urlstyle\endcsname\relax
  \providecommand{\doi}[1]{doi: #1}\else
  \providecommand{\doi}{doi: \begingroup \urlstyle{rm}\Url}\fi

\bibitem[Alian and Shelley(2014)]{alian2014photoplethysmography}
Aymen~A Alian and Kirk~H Shelley.
\newblock Photoplethysmography.
\newblock \emph{Best Practice \& Research Clinical Anaesthesiology},
  28\penalty0 (4):\penalty0 395--406, 2014.

\bibitem[Reisner et~al.(2008)Reisner, Shaltis, McCombie, Asada, Warner, and
  Warner]{REISNER2008}
Andrew Reisner, Phillip A. Shaltis, Devin McCombie, H Harry Asada, David S.
  Warner, and Mark A. Warner.
\newblock {Utility of the Photoplethysmogram in Circulatory Monitoring}.
\newblock \emph{Anesthesiology}, 108\penalty0 (5):\penalty0 950--958, 05 2008.
\newblock ISSN 0003-3022.
\newblock \doi{10.1097/ALN.0b013e31816c89e1}.

\bibitem[Charlton et~al.(2022)Charlton, Kyriacou, Mant, Marozas, Chowienczyk,
  and Alastruey]{charlton2022wearable}
Peter~H Charlton, Panicos~A Kyriacou, Jonathan Mant, Vaidotas Marozas, Phil
  Chowienczyk, and Jordi Alastruey.
\newblock Wearable photoplethysmography for cardiovascular monitoring.
\newblock \emph{Proceedings of the IEEE}, 110\penalty0 (3):\penalty0 355--381,
  2022.

\bibitem[Mejía-Mejía et~al.(2022)Mejía-Mejía, Allen, Budidha, El-Hajj,
  Kyriacou, and Charlton]{MEJIAMEJIA2022}
Elisa Mejía-Mejía, John Allen, Karthik Budidha, Chadi El-Hajj, Panicos~A.
  Kyriacou, and Peter~H. Charlton.
\newblock 4 - photoplethysmography signal processing and synthesis.
\newblock In John Allen and Panicos Kyriacou, editors,
  \emph{Photoplethysmography}, pages 69--146. Academic Press, 2022.
\newblock ISBN 978-0-12-823374-0.
\newblock \doi{10.1016/B978-0-12-823374-0.00015-3}.

\bibitem[Nitzan and Ovadia-Blechman(2022)]{NITZAN2022}
Meir Nitzan and Zehava Ovadia-Blechman.
\newblock 9 - physical and physiological interpretations of the ppg signal.
\newblock In John Allen and Panicos Kyriacou, editors,
  \emph{Photoplethysmography}, pages 319--340. Academic Press, 2022.
\newblock ISBN 978-0-12-823374-0.
\newblock \doi{10.1016/B978-0-12-823374-0.00009-8}.

\bibitem[Mukkamala et~al.(2022)Mukkamala, Hahn, and
  Chandrasekhar]{MUKKAMALA2022}
Ramakrishna Mukkamala, Jin-Oh Hahn, and Anand Chandrasekhar.
\newblock 11 - photoplethysmography in noninvasive blood pressure monitoring.
\newblock In John Allen and Panicos Kyriacou, editors,
  \emph{Photoplethysmography}, pages 359--400. Academic Press, 2022.
\newblock ISBN 978-0-12-823374-0.
\newblock \doi{10.1016/B978-0-12-823374-0.00010-4}.

\bibitem[Fine et~al.(2021)Fine, Branan, Rodriguez, Boonya-ananta, Ajmal,
  Ramella-Roman, McShane, and Coté]{fine2021}
Jesse Fine, Kimberly~L. Branan, Andres~J. Rodriguez, Tananant Boonya-ananta,
  Ajmal, Jessica~C. Ramella-Roman, Michael~J. McShane, and Gerard~L. Coté.
\newblock Sources of inaccuracy in photoplethysmography for continuous
  cardiovascular monitoring.
\newblock \emph{Biosensors}, 11\penalty0 (4), 2021.
\newblock \doi{10.3390/bios11040126}.

\bibitem[Elgendi(2012)]{elgendi2012analysis}
Mohamed Elgendi.
\newblock On the analysis of fingertip photoplethysmogram signals.
\newblock \emph{Current cardiology reviews}, 8\penalty0 (1):\penalty0 14--25,
  2012.

\bibitem[Li et~al.(2018)Li, Liu, Wu, Tang, and Li]{li2018comparison}
Suyi Li, Lijia Liu, Jiang Wu, Bingyi Tang, and Dongsheng Li.
\newblock Comparison and noise suppression of the transmitted and reflected
  photoplethysmography signals.
\newblock \emph{BioMed research international}, 2018, 2018.

\bibitem[Moscato et~al.(2022)Moscato, Palmerini, Palumbo, and
  Chiari]{MOSCATO2022A}
S.~Moscato, L.~Palmerini, P.~Palumbo, and L.~Chiari.
\newblock Quality assessment and morphological analysis of photoplethysmography
  in daily life.
\newblock \emph{Front. Digit. Health}, 4:\penalty0 912353, 2022.
\newblock \doi{10.3389/fdgth.2022.912353}.

\bibitem[Elgendi(2016)]{ELGENDI2016}
Mohamed Elgendi.
\newblock Optimal signal quality index for photoplethysmogram signals.
\newblock \emph{Bioengineering}, 3\penalty0 (4), 2016.
\newblock ISSN 2306-5354.
\newblock \doi{10.3390/bioengineering3040021}.

\bibitem[Torres-Soto and Ashley(2020)]{TORRES-SOTO2020}
Jessica Torres-Soto and Euan~A. Ashley.
\newblock Multi-task deep learning for cardiac rhythm detection in wearable
  devices.
\newblock \emph{npj Digital Medicine}, 3\penalty0 (1):\penalty0 1--8, 2020.
\newblock \doi{10.1038/s41746-020-00320-4}.

\bibitem[Mohagheghian et~al.(2022)Mohagheghian, Han, Peitzsch, Nishita, Ding,
  Dickson, DiMezza, Otabil, Noorishirazi, Scott, Lessard, Wang, Whitcomb, Tran,
  Fitzgibbons, McManus, and Chon]{MOHAGHEGHIAN2022}
Fahimeh Mohagheghian, Dong Han, Andrew Peitzsch, Nishat Nishita, Eric Ding,
  Emily~L. Dickson, Danielle DiMezza, Edith~M. Otabil, Kamran Noorishirazi,
  Jessica Scott, Darleen Lessard, Ziyue Wang, Cody Whitcomb, Khanh-Van Tran,
  Timothy~P. Fitzgibbons, David~D. McManus, and Ki~H. Chon.
\newblock Optimized signal quality assessment for photoplethysmogram signals
  using feature selection.
\newblock \emph{IEEE Transactions on Biomedical Engineering}, 69\penalty0
  (9):\penalty0 2982--2993, 2022.
\newblock \doi{10.1109/TBME.2022.3158582}.

\bibitem[Dias et~al.(2022)Dias, Costa, Cardenas, Toledo, Kriger, and
  Gutierrez]{DIAS2022}
F.~M. Dias, T.~B. Costa, D.~A.~C. Cardenas, M.~A.~F. Toledo, J.~E. Kriger, and
  M.~A. Gutierrez.
\newblock A machine learning approach to predict arterial blood pressure from
  photoplethysmography signal.
\newblock In \emph{2022 Computing in Cardiology (CinC)}, volume~XX, pages 1--4,
  2022.
\newblock \doi{XX}.

\bibitem[Goldberger et~al.(2000)Goldberger, Amaral, Glass, Hausdorff, Ivanov,
  Mark, Mietus, Moody, Peng, and Stanley]{MIMIC2}
Ary~L. Goldberger, Luis A.~N. Amaral, Leon Glass, Jeffrey~M. Hausdorff,
  Plamen~Ch. Ivanov, Roger~G. Mark, Joseph~E. Mietus, George~B. Moody,
  Chung-Kang Peng, and H.~Eugene Stanley.
\newblock Physiobank, physiotoolkit, and physionet.
\newblock \emph{Circulation}, 101\penalty0 (23):\penalty0 e215--e220, 2000.
\newblock \doi{10.1161/01.CIR.101.23.e215}.

\bibitem[Bishop and Ercole(2018)]{MQSPD}
Steven~M. Bishop and Ari Ercole.
\newblock Multi-scale peak and trough detection optimised for periodic and
  quasi-periodic neuroscience data.
\newblock In Thomas Heldt, editor, \emph{Intracranial Pressure {\&}
  Neuromonitoring XVI}, pages 189--195, Cham, 2018. Springer International
  Publishing.

\bibitem[Chen and Guestrin(2016)]{xgboost}
Tianqi Chen and Carlos Guestrin.
\newblock {XGBoost}.
\newblock In \emph{Proceedings of the 22nd {ACM} {SIGKDD} International
  Conference on Knowledge Discovery and Data Mining}. {ACM}, aug 2016.
\newblock \doi{10.1145/2939672.2939785}.

\bibitem[Prokhorenkova et~al.(2017)Prokhorenkova, Gusev, Vorobev, Dorogush, and
  Gulin]{catboost}
Liudmila Prokhorenkova, Gleb Gusev, Aleksandr Vorobev, Anna~Veronika Dorogush,
  and Andrey Gulin.
\newblock Catboost: unbiased boosting with categorical features, 2017.

\bibitem[Breiman(2001)]{randomforest}
Leo Breiman.
\newblock Random forests.
\newblock \emph{Machine Learning}, 45:\penalty0 5--32, 2001.
\newblock \doi{10.1023/A:1010933404324}.

\end{thebibliography}

\end{document}